\documentclass{article}
\usepackage{ijcai09}
\usepackage{times}
\usepackage{graphicx}
\usepackage{latexsym}
\usepackage{float}
\usepackage{epic}
\usepackage{xspace}
\usepackage{amsmath}
\usepackage{amssymb}
\input{epsf}
\floatstyle{ruled}
\restylefloat{figure}

\newtheorem{theorem}{Theorem}
\newtheorem{lemma}{Lemma}
\newtheorem{corollary}{Corollary}
\newtheorem{definition}{Definition}
\newtheorem{example}{Example}

\newcommand{\qed}{\mbox{$\Box$}}

\newcommand{\true}{\textsc{true}\xspace}
\newcommand{\false}{\textsc{false}\xspace}

\newcommand{\andgate}{\mbox{\ensuremath{\land}-gate}\xspace}
\newcommand{\andgates}{\mbox{\ensuremath{\land}-gates}\xspace}
\newcommand{\orgate}{\mbox{\ensuremath{\lor}-gate}\xspace}
\newcommand{\orgates}{\mbox{\ensuremath{\lor}-gates}\xspace}

\newlength{\halftextwidth}
\newcommand{\SLIDE}{\mbox{\sc Slide}\xspace}
\newcommand{\CARDPATH}{\mbox{\sc CardPath}\xspace}

\newcommand{\precedence}{\mbox{\sc Precedence}\xspace}

\newcommand{\sequence}{\mbox{\sc Sequence}\xspace}
\newcommand{\alldiff}{\mbox{\sc AllDifferent}\xspace}
\newcommand{\mytable}{\mbox{\sc Table}\xspace}
\newcommand{\gcc}{\mbox{\sc GCC}\xspace}

\newcommand{\among}{\mbox{\sc Among}\xspace}
\newcommand{\regular}{\mbox{\sc Regular}\xspace}
\newcommand{\roots}{\mbox{\sc Roots}\xspace}
\newcommand{\range}{\mbox{\sc Range}\xspace}
\newcommand{\grammar}{\mbox{\sc Grammar}\xspace}
\newcommand{\ctable}{\mbox{\sc Table}\xspace}

\newcommand{\bbf}{{\bf b}}
\newcommand{\Dbf}{{\bf D}}
\newcommand{\Xbf}{{\bf X}}
\newcommand{\xbf}{{\bf x}}

\newcommand{\ybf}{{\bf y}}
\newcommand{\calC}{{\mathcal{C}}}
\newcommand{\calP}{{\mathcal{P}}}

\newcommand{\notg}{\overline{g}}

\newcommand{\notp}{\overline{p}}

\newcommand{\notr}{\overline{r}}

\newcommand{\notx}{\overline{x}}
\newcommand{\noty}{\overline{y}}
\newcommand{\notz}{\overline{z}}

\newcommand{\regin}{R\'egin\xspace}

\newcommand{\set}[1]{\left\{ #1 \right\}}

\newcommand{\myOmit}[1]{}

\newcommand{\Dbfsat}{\ensuremath{{\bf D}^{sat}}}
\newcommand{\Dbfb}{\ensuremath{{\bf D}^{b}}}

\newcounter{lineno}
  {\end{alltt}}

{\makeatletter
 \gdef\xxxmark{%
   \expandafter\ifx\csname @mpargs\endcsname\relax 
     \expandafter\ifx\csname @captype\endcsname\relax 
       \marginpar{xxx}
     \else
       xxx 
     \fi
   \else
     xxx 
   \fi}
 \gdef\xxx{\@ifnextchar[\xxx@lab\xxx@nolab}
 \long\gdef\xxx@lab[#1]#2{{\bf [\xxxmark #2 ---{\sc #1}]}}
 \long\gdef\xxx@nolab#1{{\bf [\xxxmark #1]}}
}

\title{Circuit Complexity and Decompositions of Global Constraints}

\author{Christian Bessiere\thanks{Supported by the project  ANR-06-BLAN-0383-02.}\\
LIRMM, CNRS\\
Montpellier\\
bessiere@lirmm.fr
\And
George Katsirelos\thanks{NICTA is funded by the Australian Government
through the Department of Broadband, 
Communications and the Digital Economy and the Australian Research Council.}\\
NICTA \\ 
Sydney\\
gkatsi@gmail.com
\And
Nina Narodytska$\mbox{}^{\dagger}$\\
NICTA and UNSW\\
Sydney\\
ninan@cse.unsw.edu.au
\And
Toby Walsh$\mbox{}^{\dagger}$\\
NICTA and UNSW\\
Sydney\\
toby.walsh@nicta.com.au}

\newcommand{\cb}[2][CB]
{$^{\fboxsep=1pt\fbox{\tiny #1}}$%
\marginpar{\fbox{\parbox[t]{\marginparwidth}{\tiny #2}}}}

\begin{document}

\maketitle

\begin{abstract}

  We show that tools from circuit complexity can be used
  to study decompositions of global constraints. 
  In particular, we study decompositions of 
  global constraints into conjunctive normal form with the property
  that unit propagation on the decomposition enforces the same level 
  of consistency as a specialized propagation algorithm. We prove that a
  constraint propagator has a a polynomial size decomposition if
  and only if it can be computed by a polynomial size monotone Boolean
  circuit. Lower bounds on the size of monotone Boolean circuits thus 
  translate to lower bounds on the size of decompositions of global
  constraints. For instance, we prove that there is no polynomial sized 
  decomposition of the domain consistency propagator for the \alldiff 
  constraint. 
\end{abstract}

\sloppy

\section{Introduction}

Global constraints are a vital component of
constraint toolkits. They permit users to model
common patterns and
to exploit efficient propagation algorithms
to reason about these patterns. 
A promising mechanism to implement such global
constraints is to develop decompositions into 
sets of primitive constraints that do not hinder
propagation. For example, Bacchus has shown how to
decompose global propagators 
for the generic \mytable constraint, as well 
as for the \regular, \among and \sequence constraints into
conjunctive normal form (CNF) \cite{Bacchus07GAC}. 
Such decompositions can then be
used in SAT solvers, allowing
us to profit from  techniques like
clause learning and backjumping. 
In recent years, many other decompositions have been proposed
for a wide range of global constraints
including \regular and \grammar 
\cite{qwcp06,qwcp07,qwaaai08,knwcpaior08},
\sequence 
\cite{bnqswcp07},
\precedence 
\cite{wecai06},
\CARDPATH and \SLIDE \cite{slide}.
Many other global constraints 
can be decomposed using \roots and \range,
which can themselves be propagated effectively 
using some simple decompositions 
\cite{rootsrange,range,roots}. 
Finally, many global constraints specified
by automata can be decomposed into signature
and transition constraints without hindering
propagation \cite{mustadd}.

 
This raises
the important open question of which global constraints can
be effectively propagated using simple encodings \cite{BvH03}.
We show that circuit complexity can be used to resolve this question.
Our main result is that there is a polynomial sized
decomposition of a constraint propagator into CNF if
and only if the propagator
can be computed by a polynomial size monotone Boolean
circuit. It follows therefore that bounds on the 
size of monotone Boolean circuits 
give bounds on the size of decompositions of global
constraints into CNF.
For instance, a super-polynomial lower bound 
on the size of a Boolean circuit for perfect matching in
a bipartite graph gives
a super-polynomial lower bound on the size of a CNF
decomposition of the domain consistency propagator 
for the \alldiff constraint. 
Our results directly extend to decompositions into CSP constraints of
bounded arity with domains given in extension since such
decompositions can be translated into clauses of polynomial size
\cite{bhw03}.
The tools of circuit complexity 
are thus useful in understanding the limits of what we can
achieve with decompositions. 

\section{Background}

\paragraph{CSP.}

A constraint satisfaction problem (CSP) $P$ consists of a set of
variables $\Xbf$, each of which has a finite domain $D(X_i)$, and a
set of constraints $\calC$. An \emph{assignment} to a variable $X_i$
is a mapping of $X_i$ to a value $j \in D(X_i)$, 
called literal,
and written
$X_i=j$. We write
$\Dbf(\Xbf)$ 
(resp. $\Dbf'(\Xbf)$)
for sets of 
literals
$\{ X_i=j \mid X_i \in \Xbf \land
j \in D(X_i) \}$
(resp. $\{ X_i=j \mid X_i \in \Xbf \land
j \in D'(X_i) \}$)
and
$\calP(\Dbf)$ for the set
of all such sets.
An assignment to a set of variables $\Xbf$ is a set that
contains exactly one
assignment to each variable in $\Xbf$.
A constraint $C \in \calC$ has a
\emph{scope}, denoted $scope(C) \subseteq \Xbf$ and allows a subset of
the possible assignments to the variables $scope(C)$, 
called \emph{solutions} of $C$. 
A solution of $P$ is an assignment of 
one value to each variable such that
all constraints are satisfied.

%
A propagator for a constraint $C$ is an algorithm which
takes as input the domains of the variables in $scope(C)$ and returns
\emph{restrictions} of these domains. 
%
%
Following~\cite{SS04}, 
we can formally define a propagation algorithm as a function:

\begin{definition}[Propagator]
  \label{defn:propagator}
  A propagator $f$ for a constraint $C$ 
  is a
  polynomial time computable
  function $f: \calP(\Dbf) \rightarrow \calP(\Dbf)$,
  such that
  $f$ is \emph{monotone}, i.e., 
  $\Dbf'(\Xbf) \subseteq \Dbf(\Xbf) \implies 
  f(\Dbf'(\Xbf)) \subseteq f(\Dbf(\Xbf))$,
  \emph{contracting}, i.e.,
  $f(\Dbf(\Xbf)) \subseteq \Dbf(\Xbf)$,
  and 
  \emph{idempotent}, i.e., $f(f(\Dbf(\Xbf)))=f(\Dbf(\Xbf))$.
  If a literal $X_i=j$ is in $\Dbf(\Xbf) \setminus f(\Dbf(\Xbf))$ 
  then $X_i=j$ does not belong to any solution of $C$ 
  given $\Dbf(\Xbf)$. If 
  $f$ detects that
  $C$ has no solutions  under $\Dbf(\Xbf)$
  then $f(\Dbf(\Xbf))=\emptyset$.
\end{definition}

A propagator \emph{detects dis-entailment} 
if when no possible assignment is a solution of $C$
then $f(\Dbf(\Xbf)) = \emptyset$.
A propagator enforces \emph{domain consistency (DC)} when $X_i=j \in
f(\Dbf(\Xbf))$ implies that there exists a solution of $C$ that contains
$X_i=j$. 

%
We also define the \emph{consistency checker}  for a constraint $C$ 
as  
a function
that returns 0 when it detects that no
possible assignment is a solution of the  constraint and 1
otherwise, rather than restricting domains.

\begin{definition}[Consistency checker]
  \label{defn:consistency-checker}
  A consistency checker $f$ for a constraint $C$ 
  is a 
  polynomial time computable
  function 
  $f: \calP(\Dbf) \rightarrow \{0, 1\}$ 
  such that $f$ is monotone, i.e., 
  $\Dbf'(\Xbf) \subseteq \Dbf(\Xbf) \implies 
  f(\Dbf'(\Xbf)) \leq f(\Dbf(\Xbf))$.
  If $f(\Dbf(\Xbf))=0$ then no possible assignment under $\Dbf(\Xbf)$
  is a solution of $C$.
\end{definition}

We can obtain
a polynomial time 
consistency
checker $f_C$ of a constraint $C$
from a polynomial time
propagator $f_P$ for $C$ 
and vice versa~\cite{BHHW07}. 
Given the propagator $f_{P}$, the
corresponding
consistency checker
$f_C$ is defined as:
\begin{eqnarray}
  f_C(\Dbf(\Xbf)) = \left\{
      \begin{array}{ll}
        0 & f_{P}(\Dbf(\Xbf)) = \emptyset \\
        1 & otherwise
      \end{array}
      \right.
\end{eqnarray}
Conversely, given $f_C$,
the propagator $f_P$ is 
\begin{eqnarray}
  f_{P}(\Dbf(\Xbf)) = \Dbf(\Xbf) \setminus
  \left\{ X_i = j~|~ f_C(\Dbf(\Xbf)|_{X_i=j})=0 \right\}
  \label{eqn:fc2fp}
\end{eqnarray}
where $\Dbf(\Xbf)|_{X_i=j} = \Dbf(\Xbf) \setminus \left\{X_i = k|k
  \neq j\right\}$.

\paragraph{SAT.}

The Boolean satisfiability problem (SAT) is a special case of
the CSP where variables are Boolean. For each Boolean variable
$x_i$ there exist two \emph{literals} $x_i$ and
$\overline{x_i}$. Constraints in conjunctive normal
form (CNF) are disjunctions of literals,
called \emph{clauses}
and sometimes written simply as tuples of literals.

Unit propagation
\emph{forces} a literal to \true if it appears in a clause where all
other literals are \false and continues until a fix-point is reached.
If all literals in a clause are made \false, we say that the 
empty clause is produced. 
A stronger form of inference is the \emph{failed literal
  test}~\cite{Freeman-thesis}. 
For each literal $l$ of
an unset variable $x$, the failed
literal test sets $l$ to \true, 
performs unit propagation, checks whether
the empty clause was produced and retracts
$l$ and its consequences. If the empty clause
was produced, $l$ is set to \false.

A CSP instance can be encoded as a SAT instance. 
The most widely used mapping of CSP
variables to Boolean variables is the \emph{direct encoding}.
Each CSP variable  $X_i$ with domain $D(X_i)$ is
  encoded in SAT as 
    a set of propositions $x_{i,j}$, $X_i \in \Xbf, j \in D(X_i)$
    such that $X_i \neq j \iff \notx_{i,j}$.  
The  property that each CSP variable has at most one value is
    enforced by the set of clauses $(\notx_{i,j}, \notx_{i,k})$ for
    all $k \in D(X_i), k \neq j$ and the property that each CSP
    variable has at least one value is enforced by the set of
    clauses $\bigvee_{j \in D(X_i)} x_{i,j}$.
We denote this propositional representation 
of $\Dbf(\Xbf)$ as $\Dbfsat(\Xbf)$.

Note that the propositional representation $\Dbfsat(X)$ 
represents the \emph{current state} of
the domains $\Dbf(\Xbf)$ during search. This means that when the
domains change, we need to be able to make the corresponding change in
the direct encoding. Consequently, the fact $(X_i=j) \in \Dbf(\Xbf)$
is represented by $x_{i,j}$ being \emph{unset}, rather than
\true. When the value $X_i=j$ is pruned, then $x_{i,j}$ is set to
\false. Only when $X_i=j$ is the only possible assignment for $X_i$ is
$x_{i,j}$ set to \true. This means that the same domain can be
represented by different partial instantiations of the direct
encoding. For example, given the CSP variable $X_1$ with initial
domain $\set{1,2,3}$, the instantiation 
$\Dbfsat(\{X_1\}) = \{\notx_{1,2}, \notx_{1,3}\}$
(with $x_{1,1}$ unset)
corresponds to the same domain as $\Dbfsat(\{X_1\}) = \{x_{1,1},
\notx_{1,2}, \notx_{1,3}\}$, which is $\Dbf(\{X_1\}) = \set{X_1 = 1}$.

\paragraph{Boolean Circuits.}

A Boolean circuit $S$ is a directed acyclic graph (DAG). Each source
vertex of the DAG is an \emph{input gate} and the unique sink of the
DAG is the output gate. Each non-input vertex is labelled with a
logical connective, such as and ($\land$), or ($\lor$) and not
($\lnot$). An \emph{input} $\bbf$ 
to the circuit is an assignment of 
a \emph{value} 
0 or 1
to each input gate.\footnote{
  This is in contrast to \true and \false for SAT variables.}
The value of a non-input gate is
computed by applying the connective that it is labelled with to the
values of its ancestor gates. The value of the circuit $S(\bbf)$ 
is the value of
its output gate.
%

%
Any polynomial time decision algorithm can be encoded as a Boolean
circuit of polynomial size for a fixed length input~\cite{PaSt82}. 

In this paper, we will use a restriction of Boolean circuits to
\andgates and \orgates, called \emph{monotone circuits}.
The family of functions that are computable by
monotone circuits is exactly all the monotone 
Boolean
functions. 
Note that there exist families of polynomial time
computable monotone Boolean functions such that the smallest monotone
circuit that computes them is super-polynomial in
size~\cite{Razborov85}.

\begin{definition}[Monotone Boolean function]
  \label{defn:monotone-boolean-function}
  A Boolean function $f$ is monotone iff 
  $f(\bbf)=0$ implies
  $f(\bbf')=0$ for all $\bbf' \leq \bbf$, where $\leq$ is the pairwise
  vector comparison, i.e., $b_i' \leq b_i$ for all $i$.
\end{definition}

A consistency checker $f_C$,
previously defined as a monotone function over sets,
can also be formalised as a monotone Boolean
function whose input is the characteristic function of the set
$\Dbf(\Xbf)$.  Literals $X_i=j$ are mapped to arguments
$b_{i,j}$ of the function, with 
$b_{i,j}=1$ iff $X_i=j \in \Dbf(\Xbf)$. We use $\Dbfb(\Xbf)$ to
denote the setting of the $b_{i,j}$ inputs for a given set of domains
$\Dbf(\Xbf)$.


\section{Properties of CNF decompositions}
\label{sec:decompositions}

In this section, we define formally a 
\emph{CNF decomposition} of a propagator and of a consistency
checker. As with propagators and consistency
checkers~\cite{BHHW07}, we show that there exists a polynomial time
conversion between the CNF decompositions of a propagator and of the
corresponding consistency checker.

\begin{definition}[CNF Decomposition of a propagator]
  \label{defn:decomp-prop}
  A CNF decomposition of a propagation algorithm $f_P$ 
  is a 
  formula in CNF $C_P$ over
  variables $\xbf \cup \ybf$ such that
  \begin{itemize}
  \item The \emph{input variables} $\xbf$
    are the propositional representation
    $\Dbfsat(\Xbf)$ of $\Dbf(\Xbf)$ and
    $\ybf$ is a set of auxiliary variables whose size
    is polynomial in
    $|\xbf|$.
  \item $x_{i,j}$ is set to \false by unit propagation if and only if
    $X_i=j \notin f_P(\Dbf(\Xbf))$.
  \item Unit propagation on $C_P$ produces the empty clause when
    $f_P(\Dbf(\Xbf)) = \emptyset$.
  \end{itemize}
\end{definition}

\begin{example}
\label{exm:def_prop}
To illustrate Definition~\ref{defn:decomp-prop}, consider 
a \ctable constraint over the variables $X_1, X_2$ with
$D(X_1) = D(X_2) = \{a,b\}$ and
the satisfying assignments: $\left\{\left\langle a,a\right\rangle, 
\left\langle
  b,b\right\rangle\, \left\langle a,b \right\rangle \right\}$. 
\cite{Bacchus07GAC} decomposes such a \ctable constraint into CNF
using the following set of clauses:
$$
\begin{array}{cc c cc }
x_{1a} \Rightarrow y_1 \lor y_3 &  x_{2a} \Rightarrow y_1 & y_{1} \Rightarrow x_{1a} &  y_1 \Rightarrow x_{2a} &  \\ 
x_{1b} \Rightarrow y_2 &  x_{2b} \Rightarrow y_2 \lor y_3 &  y_{2} \Rightarrow x_{1b} &  y_2 \Rightarrow x_{2b}& \\
y_3 \Rightarrow x_{1a} & y_3 \Rightarrow x_{2b} &
y_1 \vee y_2 \vee y_3
\end{array}
$$
where $\xbf = \{x_{i,j}\}$, $i \in \{1,2\}$, $j \in \{a,b\}$ is the
propositional representation $\Dbfsat(\Xbf)$ 
of $\Dbf(\Xbf)$ and $\ybf = \{y_i\}$, $i \in
\{1,2,3\}$ are auxiliary variables that correspond to satisfying
tuples. 
Note that we have extended Bacchus's encoding with the clause $(y_1
\lor y_2 \lor y_3)$ to detect failure.
Suppose the value $a$ is removed from the domain of $X_1$. The
assignment $x_{1a} = \false$ forces the variable $y_1$ to \false,
which in turn causes the variable $x_{2a}$ to \false, removing the
value $a$ from the domain of $X_2$ as well.
\end{example} 

In example~\ref{exm:def_prop}, we have decomposed a
constraint into clauses
by introducing variables. In
general, an encoding might be exponentially bigger if auxiliary
variables are not used (e.g., the parity function~\cite{DM02}). 

\begin{definition}\emph{\textbf{(CNF Decomposition of a 
      consistency checker)}}
  \label{defn:cnf-decomposition}
  A CNF decomposition of a consistency checker $f_C$ is a CNF $C_C$ over
  variables $\xbf \cup \ybf \cup \{ z \}$ such that
  \begin{itemize}
  \item The input variables $\xbf$
    are the propositional representation
    $\Dbfsat(\Xbf)$ of $\Dbf(\Xbf)$
    and
    $\ybf$ is a set of auxiliary variables whose size
    is polynomial in
    $|\xbf|$. The variable $z$ is the \emph{output variable}.
  \item Unit propagation
    on $C_C$ never forces any variable from $\xbf$
    or generates the empty clause
    if no variable in 
    $\ybf$ is set externally to $C_C$, i.e.,
    every variable $y\in \ybf$ is either unset or 
    forced by a clause in $C_C$. 
  \item $z$ is set to \false by unit propagation if and only if 
    $f_C(\Dbf(\Xbf)) = 0$.
  \end{itemize}
\end{definition}

\begin{example}
\label{exm:def_con_checker}
Consider the \ctable constraint from Example~\ref{exm:def_prop}. We
construct a CNF decomposition of a consistency checker using the CNF
decomposition of a propagator.  The clauses that cause pruning of
input variables domains are removed and the last clause is
augmented with the output variable $z$
to avoid generation of the
empty clause in the case of failure:
$$
\begin{array}{cc c cc}
y_{1} \Rightarrow x_{1a} &  y_1 \Rightarrow x_{2a}  &
y_{2} \Rightarrow x_{1b} &  y_2 \Rightarrow x_{2b} \\
y_{3} \Rightarrow x_{1a} & y_3 \Rightarrow x_{2b} &
\noty_1 \land \noty_2 \land \noty_3 \Rightarrow \notz
\end{array}
$$
In this case, if the value $a$ is removed from the domain of $X_1$,
unit propagation will not deduce that $a$ has to be removed from the
domain of $X_2$.
Consider instead
the case when the values $a$ and $b$ are removed from
the domains of $X_1$ and $X_2$, respectively.  The literals $x_{1a} =
\false$ and $x_{2b} = \false$ force the auxiliary variables $y_1$, 
$y_2$ and $y_3$
to be \false.  Therefore, the output variable $z$ is forced to
\false, signalling that the \ctable constraint does not have a solution
under $\Dbf(\Xbf)$.
\end{example} 

In example~\ref{exm:def_con_checker}, we transformed the propagator of
example~\ref{exm:def_prop} into a consistency checker in an ad-hoc
manner. The next theorem shows
that this can be done in a generic way.
We give
a polynomial transformation of
CNF decompositions of a propagator 
into consistency checkers 
This mirrors the results of~\cite{BHHW07} 
for CNF decompositions.

\begin{theorem}
  \label{thm:flt_transformation}
  There exists a polynomial time and space conversion between the CNF
  decomposition of a propagator $f_P$ and that of the corresponding
  consistency checker $f_C$.

\end{theorem}

\begin{proof}
  $(\rightarrow)$
  We construct $C_C$ as a transformation of $C_P$ such that  the
  output variable $z$ of $C_C$ is \false iff unit propagation
  on $C_P$ produces the empty clause. 

  Let the set of clauses 
  of $C_P$
  be $c_1 \ldots c_m$. 
  For each variable $p \in \xbf \cup \ybf$,
  we introduce 2 variables $p_t$ and $p_f$ in $C_C$
  so that $p_{t}$ and $p_{f}$ are true
  if $p$ is forced to \true or \false, respectively:
  \begin{eqnarray}
    p \implies p_{t} &&
    \notp \implies p_{f} \label{eqn:up_reify_channel}
  \end{eqnarray}
  
  Then, we simulate unit propagation 
  for each clause $c_k$ by replacing it
  with 3 implications\footnote{
    We assume that formulas are given in 3-CNF
    form. We can 
    convert any CNF formula to 3-CNF, increasing
    its size by at most a constant factor
    and without hindering unit 
    propagation~\cite[section
    3.1.1]{GareyJohnsonBook}.}
  that contain the variables $p_t$ and $p_f$ rather than $p$.
  For example, to simulate unit
  propagation for the clause $c_1 = (p,q,\notr)$, we 
  replace it with 
  \begin{eqnarray}
    p_{f} \land q_{f} \implies r_{f} &
    p_{f} \land r_{t} \implies q_{t} &
    q_{f} \land r_{t} \implies p_{t} \label{eqn:up_reify_sim}
  \end{eqnarray}

  Unit propagation on \eqref{eqn:up_reify_sim} 
  can never derive the empty clause, because
  the true and false
  values of $p$ are encoded in different variables
  $p_{t}$ and $p_{f}$, which may be true simultaneously. 
  When this happens, unit propagation on $C_P$ would
  generate the empty clause,
  therefore we must set the output variable $z$ to \false,
  using the following clauses:
  \begin{eqnarray}
    p_t \land p_f \implies \notz \label{eqn:up_reify_output}
  \end{eqnarray}

  The union of the
  clauses \eqref{eqn:up_reify_channel}, \eqref{eqn:up_reify_sim}
  and \eqref{eqn:up_reify_output} 
is a 
  CNF decomposition of $f_C$ with size
  $O(|\xbf \cup \ybf| + |C_P|) = O(|C_P|)$,
  therefore the transformation is polynomial.

  $(\leftarrow)$
  We outline the proof here.
  We replicate the equation~\eqref{eqn:fc2fp}
  by simulating the
  failed literal test
  on $C_C \cup \{(z)\}$. 
  For each literal $x_{i,j}$ we create a copy of
  $C_C$, denoted by $C_C|_{x_{i,j}}$, 
  in which all
  literals $x_{i,k}, k \neq j$ are \false.
  We use $C_C|_{x_{i,j}}$ to record the results of unit
  propagation when $X_i=j$.
  When unit propagation sets the
  output variable $z_{x_{i,j}}$ of the copy $C_C|_{x_{i,j}}$ to \false 
  then the propositional literal $x_{i,j}$ is
  made \false by the additional clause $(\notz_{x_{i,j}} \implies \notx_{i,j})$.

  The decomposition $C_P$ is then the union of the copies of $C_C$
  and the clauses $(\notz_{x_{i,j}} \implies \notx_{i,j})$:
  \begin{eqnarray}
    C_P &=& \bigcup_{x_{i,j} \in \xbf} (C_C|_{x_{i,j}} \cup (z_{x_{i,j}}, \notx_{i,j}) )
    \label{eq:flt}
  \end{eqnarray}
  The size of $C_P$ is $O(|\xbf|\cdot|C_C|)$, therefore the transformation is
  polynomial.
  \qed
\end{proof}


Using the encoding of theorem~\ref{thm:flt_transformation},
a CNF decomposition 
of a 
consistency checker
that 
detects dis-entailment 
can
be made into a propagator that enforces
domain consistency. 
As an example, consider the CNF decomposition of a propagator that
detects dis-entailment for the \sequence constraint, proposed
in~\cite{Bacchus07GAC}. The size of this decomposition is $O(n^2)$,
where $n$ is the number of variables
in the \sequence constraint.
These variables
are binary, hence the transformation of
theorem~\ref{thm:flt_transformation} yields a decomposition of a DC
propagator with size $O(n^3)$. This is also the complexity of the DC
propagator proposed in \cite{VHPRS06}.

Since all definitions of CNF decompositions that we introduced in this
section are polynomially equivalent, in the remainder of this paper we
only prove results for CNF decompositions of consistency checkers.

\section{Equivalence to monotone circuits}

In this section, we show our main result, which establishes a
connection between CNF decompositions of constraints and circuit
complexity. 

\begin{theorem}
  \label{thm:cnf_mc_equiv}
  A consistency checker $f_C$ can be decomposed to a CNF of polynomial
  size if and only if it can be computed by a monotone circuit of
  polynomial size.
\end{theorem}

The proof of theorem~\ref{thm:cnf_mc_equiv} is constructive. We will
first show the reverse direction, using the Tseitin
encoding~\cite{Tseitin70} of a monotone circuit.

\begin{definition}[Tseitin encoding of a Boolean circuit]
  The Tseitin encoding of a circuit $S$ into clausal form has one
  propositional variable for each input of $S$ and for each gate of
  $S$. W.l.o.g, we assume all gates have fan-in 2. For each \andgate
  $g$ with inputs $x_1$, $x_2$, the Tseitin encoding contains the
  clauses $(x_1, \notg)$, $(x_2, \notg)$, $(\notx_1, \notx_2, g)$ and
  for each \orgate it contains the clauses $(\notx_1, g)$, $(\notx_2,
  g)$, $(x_1, x_2, \notg)$. Given any complete instantiation of the
  input variables, unit propagation on the Tseitin encoding sets the
  variable corresponding to the output gate of $S$ to \true if the
  circuit computes 1 and to \false otherwise.
\end{definition}

Suppose that a consistency checker
$f_C$ can be encoded into a monotone circuit $S_C$ of polynomial
size. The Tseitin encoding of $S_C$ turns out to be a
CNF decomposition of $f_C$. 
This is a direct consequence of the following lemma.

\begin{lemma}
  \label{lemma:mc_cnf}
  Let $S_C$ be a monotone circuit and $C_C$ be its Tseitin
  encoding. Let $I$ 
  be a partial instantiation of the input variables $\xbf$
  of $C_C$ and $\bbf$ be 
  the corresponding input to $S_C$, where
  $b_i = 0$ iff $\notx_i \in I$.
  Then, unit propagation on $C_C$ with $I$ forces the output
  variable $z$ to \false if and only if $S_C(\bbf)=0$.
\end{lemma}

\begin{proof}
  $(\rightarrow)$ This follows from the correctness of the Tseitin
  encoding. 

  $(\leftarrow)$. Suppose that
  $S_C(\bbf)=0$, but the output variable
  $z$ 
  is not forced to
  \false by unit propagation under $I$. 
  Consider an instantiation $I'$ of the input variables
  of $C_C$, which is the same as $I$
  with unset variables fixed to \true.
  Let $y \in \ybf
  \cup \{ z\}$ 
  be an auxiliary variable 
  that is unset under $I$. All such variables correspond
  to a gate in $S_C$.
  Since $C_C$ is an encoding of the monotone circuit $S_C$, $y$ will
  be set to \true under $I'$. 
  This means that the output variable $z$ is
  also set to \true.  By the correctness of the Tseitin encoding,
  $S_C(\bbf)=1$, a contradiction.
  \qed
\end{proof}

\begin{corollary}
  Let $S_C$ be a monotone circuit and $C_C$ be its Tseitin
  encoding. Let $I$ be a partial instantiation of the input variables
  $\xbf$ of $C_C$. Then,  
  unit propagation on $C_C$ with $I$ forces the
  output variable $z$ to \false if and only if $S_C(\bbf)=0$, 
  for all $\bbf$
  where $\bbf$ is the input to $S_C$ that corresponds 
  to any extension of $I$ to a complete instantiation.
\end{corollary}

\begin{proof}
  This follows from lemma~\ref{lemma:mc_cnf} and the fact that $S_C$
  is a monotone circuit.  \qed
\end{proof}

Interestingly, lemma~\ref{lemma:mc_cnf}
cannot be generalised to non-monotone Boolean
circuits. The next example shows that
there exists
a non-monotone 
Boolean circuit $S$ that computes a monotone function,
and a partial instantiation $I$ with
$\bbf$ the corresponding input to $S$,
such that 
$S(\bbf)=0$ 
but unit propagation on the
Tseitin encoding of $S$ under the instantiation $I$ 
does not set the output variable to \false.

\begin{figure}[htb]
  \centering
    \includegraphics[width=0.4\textwidth]{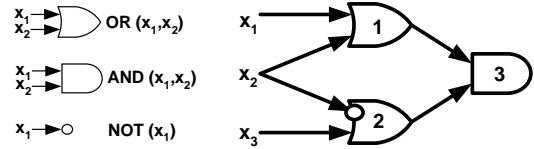}\\
  \caption{A circuit whose Tseitin encoding is incomplete.}
  \label{fig:bad_circuit}
\end{figure}

\begin{example}
  Consider the 
  non-monotone
  circuit $S$ shown in figure~\ref{fig:bad_circuit}. 
  Note that $S$ computes a monotone function.

  The Tseitin encoding of $S$ introduces three Boolean variables
  $g_1$, $g_2$ and $g_3$ for the gates $OR_1$, $OR_2$ and $AND_3$,
  respectively, and the clauses
  $(\notx_1, g_1)$, $(\notx_2, g_1)$,
  $(\notg_1, x_1, x_2)$, 
  $(\notx_1, g_2)$, $(x_2, g_2)$, $(\notg_2, x_1, \notx_2)$,
  $(\notg_3, g_1)$, $(\notg_3, g_2)$, $(\notg_1, \notg_2, g_3)$.

  Now suppose that $I=\{ \notx_1 \}$. 
  Then, $\bbf=\{x_1=0, x_2=1\}$
  and $S(\bbf)=0$. Since $S$ computes a monotone function, all
  possible extensions of $\xbf$ evaluate to 0. But in the Tseitin
  encoding, setting $x_1$ to \false does not make any clauses
  unit, therefore unit propagation does not set $g_3$ to \false.
  \qed
\end{example}

We now show the forward direction
of theorem~\ref{thm:cnf_mc_equiv}:
every CNF decomposition $C_C$
of a consistency checker $f_C$
can
be converted to a monotone circuit
that computes $f_C$
with at most a polynomial increase in size.

This transformation exploits two properties 
of CNF decompositions,
namely, that only positive literals
of input variables appear in $C_C$, and that
unit propagation
only makes auxiliary variables \false.
We show the former property in lemma~\ref{lemma:input var polarity}
and the latter in lemma ~\ref{lemma:extra_var_values}.

\begin{lemma}
  \label{lemma:input var polarity}
  Let $C_C$ be the CNF decomposition of a consistency checker
  $f_C$. There exists a polynomial size CNF decomposition $C_C'$ of
  $f_C$ such that negative literals of the input variables do not
  appear in any clause in $C_C'$.
\end{lemma}

\begin{proof}
  We construct $C_C'$ by removing from $C_C$ all clauses that contain
  a negative literal of an input variable. We show by contradiction
  that unit propagation on $C_C'$ and $C_C$ produces identical
  results for the output variable $z$.

  Let $I$ be a partial instantiation of the input variables such that
  unit propagation on $C_C$ under $I$ sets $z$ to \false but leaves
  $z$ unset on $C_C'$.
  Since unit propagation on $C_C$ and $C_C'$ produces different
  results, at least one of the removed clauses becomes unit under $I$
  in $C_C$. By definition, $C_C$ never forces any literal of an input
  variable, so for any removed clause to become unit, all the literals
  of input variables in it have to be \false. Since at least one of
  these literals is negative, at least one input variable has to be
  set to \true in $I$. 

  We construct another partial instantiation $I'$ from $I$ by setting
  the same literals to \false as $I$ and leaving the rest unset, i.e.,
  $I'=\{\notx_{i,j}| \notx_{i,j} \in I\}$. The partial instantiations
  $I$ and $I'$ represent the same domains $\Dbf(\Xbf)$, because the
  mapping from partial instantiation to domain depends only on the
  literals that are \false. By this and the fact that $C_C$ is a
  decomposition of $f_C$, unit propagation on $C_C$ under $I'$ forces
  the output variable $z$ to the same value as under $I$, \false.

  Consider the result of unit propagation on $C_C'$ under $I'$.
  Recall that by definition $C_C$ does not modify input variables and
  $I'$ does not have literal set to \true by construction.  Hence,
  none of the clauses that we remove from $C_C$ to get $C_C'$ can
  become unit after performing UP on $C_C$ under $I'$.  Hence, unit
  propagation in $C_C'$ under $I'$ sets $z$ to \false as in $C_C$.
  On the other hand, $I$ sets a superset of the literals that $I'$
  sets, so unit propagation on $C_C'$ under $I$ also sets
  $z$ to \false, a contradiction, since we assumed that $C_C'$ leaves
  $z$ unset under $I$.
\qed  
\end{proof}

In practice, a CNF decomposition of a consistency checker may not be self
contained and may depend on the existence of clauses in the direct encoding of
variable domains.  In this case, we cannot just remove clauses that
contain negative literals of input variables, 
as lemma~\ref{lemma:input var polarity} suggests. 
However, using the
clauses of the direct encoding, we can substitute negative literals
with the disjunction of positive literals. For instance, consider a
variable $X_2$ with the domain $\{ 1,2,3 \}$ and a clause $(x_{1,1},
\notx_{2,2}, \noty)$ in $C_C$.  The literal $x_{2,2}$ can make
this clause unit.  The direct encoding of $D(X_2)$ includes a clause
$(x_{2,1},x_{2,2}, x_{2,3})$.  Note that the literal $x_{2,2}$
is \true if and only if literals $x_{2,1}$ and $x_{2,3}$ are \false.
Therefore, the literal $\notx_{2,2}$ can be replaced with the
disjunction $(x_{2,1}, x_{2,3})$ and the clause $(x_{1,1},
\notx_{2,2}, \noty)$ is transformed to the clause $(x_{1,1}, x_{2,1},
x_{2,3}, \noty)$.

The next step is to show that we can transform a CNF decomposition so
that each auxiliary variable is unset or \false for \emph{all} inputs
that make the output variable \false. The transformation is a renaming
of the auxiliary variables. Lemma~\ref{lemma:extra_var_values}
describes the property that 
allows 
this transformation.

\begin{lemma} \label{lemma:extra_var_values} 
  Let $C_C$ be a CNF decomposition of a consistency checker $f_C$ over
  the variables $\xbf \cup \ybf \cup \{z\}$, 
  $I_1 = \Dbfsat_1(\Xbf), I_2 = \Dbfsat_2(\Xbf)$ 
  be the propositional representations of any two domain settings
  such that unit propagation on $C_C$
  forces $z$ to \false under both $I_1$ and $I_2$. For any variable $y
  \in \ybf$, if $y$ is forced to \false (\true) by unit propagation under
  $I_1$ then it is not forced to \true (\false) by unit propagation under
  $I_2$.
\end{lemma}

\begin{proof}
  Let a variable $y$ be forced to \true by unit propagation under
  $I_1$ and to \false under $I_2$, but $z$ is \false under both $I_1$
  and $I_2$.
  Consider the partial instantiation $I$ such that if a variable $x
  \in \xbf$ is \false in either $I_1$ or $I_2$, it is also \false in
  $I$, otherwise it is unset. 
  Since $I$ fixes a superset of the literals that are fixed in either
  $I_1$ or $I_2$, all clauses that became unit by either $I_1$ or
  $I_2$ will also be unit in $I$. Therefore, unit propagation under
  $I$ will force at least the union of the sets of literals forced by
  $I_1$ and $I_2$. This means that unit propagation under $I$ will
  make both $y$ and $\noty$ \true, which generates the empty
  clause. This is a contradiction, as $C_C$ can never produce the
  empty clause.
  %
  \qed
\end{proof}

\begin{corollary}
  \label{cor:extra_var_false}
  A CNF decomposition $C_C$ of a consistency checker $f_C$ over variables
  $\xbf \cup \ybf \cup \left\{z\right\}$, 
  can be polynomially converted into a
  decomposition $C_C'$ of $f_C$ such that every variable in $\ybf$ is
  either unset or \false when $z$ is \false.
\end{corollary}

\begin{proof}
  We construct $C_C'$ from $C_C$ by flipping the polarity of those
  variables that are set to \true when $z$ is \false. \qed
\end{proof}

Lemma~\ref{lemma:input var polarity} and
corollary~\ref{cor:extra_var_false} allow us to precisely characterize
the form of the clauses in a CNF decomposition.

\begin{corollary}
  \label{lemma:anti-horn}
  Let $C_C$ be a CNF decomposition of a consistency checker $f_C$.
  The variables of $C_C$ can be renamed so that each clause has exactly
  one negative literal.
\end{corollary}

\begin{proof}
  By lemma~\ref{lemma:input var polarity}, all input variables are
  positive literals in the decomposition
  and by definition \ref{defn:cnf-decomposition} 
  they are never forced by unit propagation on $C_C$.
  In addition, by corollary~\ref{cor:extra_var_false}, we can rename
  the auxiliary variables so that unit propagation on $C_C$ may only
  ever set them to \false.
  Then, in any clause that consists of input variables and one
  auxiliary variable $y$, $y$ must be negative, otherwise it may be
  set to \true, a contradiction.
  
  Suppose there exists a clause $c$ with two auxiliary variables $y_1$
  and $y_2$ and both are negative in $c$. Since neither $y_1$ nor
  $y_2$ can ever be made \true, this clause can never become unit and
  can be ignored. Suppose the literals of both $y_1$ and $y_2$ are
  positive in $c$. Then, if $c$ becomes unit, it makes one of the
  auxiliary variables \true, a contradiction. Thus, exactly one of the
  literals of $y_1$ and $y_2$ is negative in $c$.
  The same reasoning can be extended to clauses with more than two
  auxiliary variables.
  \qed
\end{proof}

The condition described by corollary~\ref{lemma:anti-horn} 
is similar
to $C_C$ being re-nameable anti-Horn, but is stronger
as it requires \emph{exactly} one
negative literal in each clause, rather than at most one.
This condition allows us to build a monotone circuit
from a decomposition, using the construction of the
next lemma.


\begin{lemma}
  \label{lemma:cnf-to-mc}
  Let $C_C$ be a CNF decomposition of a consistency checker $f_C$.
  Then, there exists a monotone circuit $S_C$ of size
  $O(n|C_C|)$ that computes $f_C$.
\end{lemma}

\begin{figure*}[htb]
  \centering
    \includegraphics[width=0.7\textwidth]{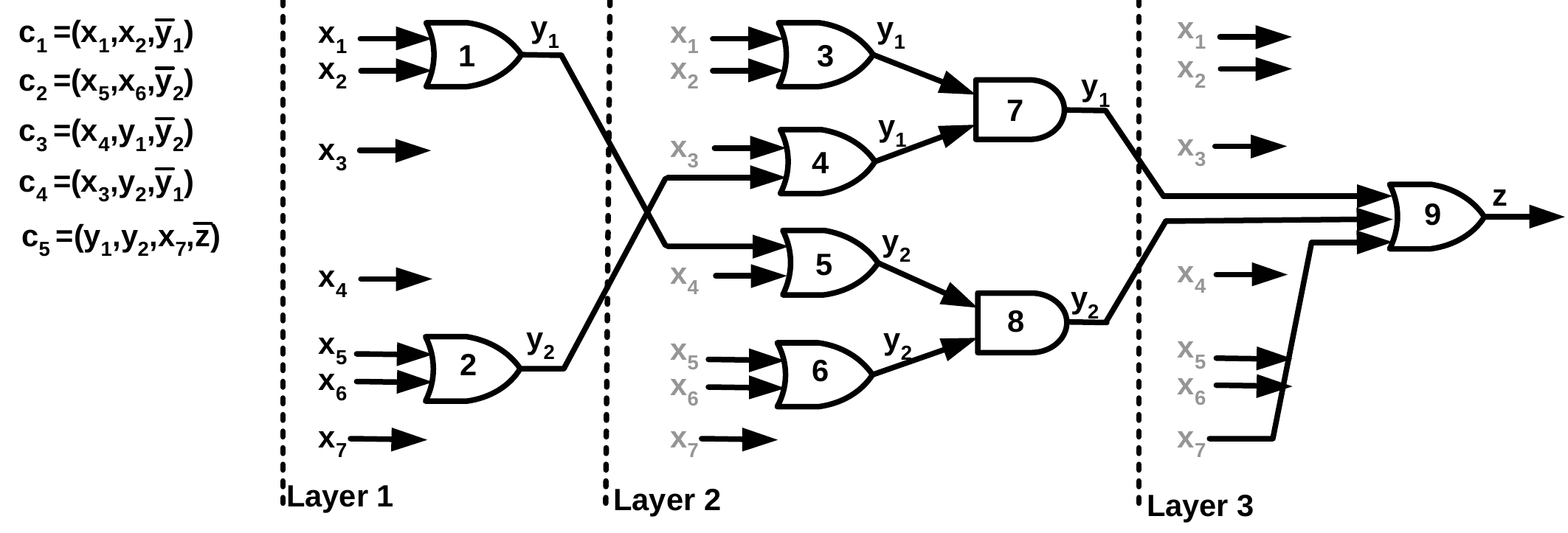}\\
  \caption{Conversion of a CNF decomposition of a consistency 
    checker into a monotone Boolean circuit.}
  \label{fig:good_circuit}
\end{figure*}

\begin{proof}
  We assume that $C_C$ is in the form described in
  corollary~\ref{lemma:anti-horn}.

  The inputs of the circuit correspond to the input variables of
  $C_C$. For each input 
  variable $x_{i,j}$ of $C_C$, there exists an input $b_{i,j}$
  of $S_C$ which is 0 if $x_{i,j}$ is \false and 1 otherwise. Internal
  gates of the circuit correspond to auxiliary variables after a
  certain number of unit propagation steps, using the same mapping.

  We create a circuit with $|\ybf|$ \emph{layers} $1 \ldots |\ybf|$. Let
  $c_1,\ldots, c_m$ be the clauses of $C_C$. 
  The $i^{th}$ layer of the circuit contains an
  \orgate $c^i_j$ for each clause $c_j$, called \emph{clause gates}
  and an \andgate $y^i_k$ for each auxiliary variable $y_k$, called
  \emph{variable gates}. Consider a clause $c_j$ which contains
  $\noty$ as the sole negative literal (recall that
  corollary~\ref{lemma:anti-horn} ensures that this is the case), the
  positive literals of input variables $x_{j_1},\ldots,x_{j_q}$ and
  the positive literals of auxiliary variables $y_{j_{q+1}}, \ldots,
  y_{j_{q+r}}$. The inputs of each gate $c_j^i$ are $b_{j_1},\ldots,
  b_{j_q}$ and $y^{i-1}_{j_{q+1}},\ldots, y^{i-1}_{j_{q+r}}$. Let the
  clauses with $\noty_k$ as the sole negative literal be $c_{k_1},
  \ldots, c_{k_s}$. Then, the inputs of each gate $y^i_k$ are
  $c^i_{k_1},\ldots, c^i_{k_s}$. The output of the circuit is
  $z^{|\ybf|}$. Note that in this construction the inputs of some the gates
  may not be defined. This is the case, for example, for the gate
  $c^1_i$, where the clause $c_i$ contains the positive literals of
  some auxiliary variables. If this happens for a clause gate, we omit
  it, while if it happens for a variable gate, we omit the undefined
  input. If all the inputs of a variable gate are undefined, we omit
  the gate.

  This construction computes one breadth first application of unit
  propagation at each layer. Specifically, the gate $y^i_k$ is 0
  iff $y_k$ is forced to \false after $i$ or fewer
  breadth first steps of unit
  propagation, while the gate $c^i_j$ is 0 iff the negated variable
  in $c_j$ is forced to \false after $i$ or fewer 
  breadth first steps of unit
  propagation.
  We show this by induction. For the first
  layer, there exist gates only for clauses with no positive literals
  of auxiliary variables. Consider any such gate $c_j$ which contains
  the negative literal $\noty_k$. All the propositional
  variables in $c_j$ except $y_k$ are \false 
  iff the corresponding inputs are 0.
  Thus $c^1_j$ is 0 iff $y_k$ is \false after
  unit propagation of $c_j$. If many clauses contain the negative
  literal $\noty_k$, then 
  at least one of them sets $y_k$ to \false in one breadth
  first step iff there exists a clause gate that is 0 and is an
  input to the variable gate $y^1_k$, which is an \andgate and is thus
   0. For the inductive step, assume that the layers $1\ldots k-1$
  compute $k-1$ breadth first steps of unit propagation. The same
  reasoning as for the base case shows that the results of unit
  propagation are correctly computed for the $k^{th}$ layer. Note that
  the $k^{th}$ layer may also contain gates that were omitted at
  previous levels. Since the inputs of these gates
  are correctly computed by the
  inductive hypothesis, the gates that are new to the $k^{th}$ layer
  are also correctly computed.

  To conclude the proof, observe that in the extreme case, unit
  propagation will set one more literal at every breadth first step,
  thus after $|\ybf|$ steps it must either arrive at a fixpoint or
  set all literals. Since the circuit has $|\ybf|$ layers, it will
  correctly compute the result of unit propagation on $C_C$.
  \qed
\end{proof}

We illustrate the construction of lemma~\ref{lemma:cnf-to-mc} with an
example.

\begin{example}
  \label{ex:cnf-to-mc}
  Consider the CNF decomposition $C_C = \{ c_1, c_2, c_3, c_4, c_5 \}$,
  where $c_1 = (x_1, x_2, \noty_1)$, $c_2 = (x_5, x_6, \noty_2)$, $c_3
  = (x_4, y_1, \noty_2)$, $c_4 = (x_3, y_2, \noty_1)$, $c_5 = (y_1, y_2,
  x_7, \notz)$.

  We construct a monotone circuit $S_C$ from $C_C$, 
  (figure~\ref{fig:good_circuit}). For a given instantiation of the
  input variables, 
  this circuit computes 0 for the corresponding Boolean inputs
  if and only if unit
  propagation on $C_C$ forces the output variable to \false.

  The circuit consists of 3 layers, with gates 1 and 2 in the first
  layer, 3--8 in the second and gate 9 in the third. The gates 1--6
  and 9 are clause gates, while gates 7 and 8 are variable gates. A
  strict application of the construction of
  lemma~\ref{lemma:cnf-to-mc} would also have variable gates in layers
  1 and 3, but we omit them here as they would be single-input gates.
  Note 
  that in figure~\ref{fig:good_circuit}, inputs are replicated at
  each layer to reduce clutter.

  We note also that the layered construction 
  of lemma~\ref{lemma:cnf-to-mc}
  is necessary.
  A circuit that attempts to
  capture unit propagation on all clauses without using layers
  would have to contain a cycle
  between the gates that compute $y_1$ and $y_2$, because $y_1$ would
  need to be an input of the clause gate $c_3$ that computes $y_2$ 
  and $y_2$ would need to an input of the clause gate $c_4$ 
  that computes
  $y_1$. Constructing a layered circuit allows us to 
  remove such cycles.
  %
  %
  \qed
\end{example}

The proof of theorem~\ref{thm:cnf_mc_equiv} is now immediate from
lemmas~\ref{lemma:mc_cnf} and \ref{lemma:cnf-to-mc}.
Since CNF decompositions of consistency checkers can be
converted in polynomial time to and from CNF decompositions of
propagators, theorem~\ref{thm:cnf_mc_equiv} also holds for propagators.

\section{Non decomposable global constraints}
\label{sec:non-decomp-some}

Corollary~\ref{cor:alldiff} now uses an existing circuit complexity result
to show that, unsurprisingly, there is no polynomial
size CNF decomposition of the domain consistency propagator for the
\alldiff constraint. This also applies to
generalizations of \alldiff, such as \gcc.

\begin{corollary}
  \label{cor:alldiff}
  There is no polynomial sized CNF decomposition of the \alldiff
  domain consistency propagator.
\end{corollary}

\begin{proof}
  \regin~\cite{Regin94} showed that an \alldiff constraint has a
  solution iff the corresponding bipartite value graph (i.e., the
  graph where the node representing a variable has an edge to every
  node that represents a value in its domain)  has a perfect
  matching. 
In addition, every bipartite graph corresponds to the value graph of
an \alldiff constraint and DC propagators detect dis-entailment. Thus,
if there exists a polynomial size CNF decomposition of the
  \alldiff DC propagator, we can construct a monotone circuit that
  computes whether a bipartite graph has a 
perfect matching. But
  Razborov~\cite{Razborov85} showed that the smallest monotone circuit
  that computes whether there exists a 
perfect matching for a
  bipartite graph is super-polynomial in the number of vertices in the
  graph. Therefore, the smallest CNF decomposition of the \alldiff
  DC propagator is super-polynomial in size. \qed
\end{proof}

On the other hand, bound and range consistency
propagators of \alldiff can be decomposed,
as we argue in~\cite{bknqw09blind}.

\section{Conclusions and Future Work}
\label{sec:conclusions}

In this paper we have shown how 
the tools of circuit complexity can be used
to study decompositions of global propagators into CNF.
%
Our  results directly extend to decompositions into CSP constraints of
bounded arity with domains given in extension since
such decompositions can be translated into clauses of polynomial size.
%
An interesting next step is to consider
the decomposability of
constraint propagators into  more expressive primitive constraints
where  domains are represented  in logarithmic space via their
bounds. CSP solvers provide this feature which is
missing in CNF. We conjecture that there exists an
equivalence between such CSP decompositions of constraint propagators and
monotone arithmetic circuits that are generalizations of Boolean
monotone circuits to real numbers and gates for addition
and multiplication. 
Since lower bound results on monotone circuits usually transfer to
monotone arithmetic circuits, this would imply
that the domain consistency propagator for \alldiff 
cannot be decomposed to constraints that exploit
(exponentially) large domains. 

\bibliographystyle{named}

\end{document}